\documentclass[conference]{IEEEtran}
\newlength \mycolumnwidth
\usepackage[mathscr]{euscript}
\usepackage{tabularx}
\usepackage{amssymb}
\usepackage{amsmath}
\usepackage{mathtools,xparse}
\DeclarePairedDelimiter{\norm}{\lVert}{\rVert}
\usepackage{graphicx}
\usepackage{graphics}
\usepackage{xcolor}
\usepackage{supertabular}
\usepackage{caption}
\usepackage{subcaption}
\usepackage{amsmath}
\usepackage[ruled, linesnumbered, vlined]{algorithm2e}

\usepackage{algpseudocode}
\usepackage{todonotes}
\DeclareMathOperator*{\EE}{\mathbb{E}}
\newcommand\E[2]{\EE_{#1}\left[#2\right]}
\DeclareMathOperator*{\argmax}{arg\,max}

\DeclarePairedDelimiter\floor{\lfloor}{\rfloor}

\usepackage{mathtools}
\usepackage{booktabs}
\renewcommand{\vec}[1]{\boldsymbol{#1}}
\renewcommand{\gets}{\leftarrow}
\let\oldnl\nl
\newcommand{\nonl}{\renewcommand{\nl}{\let\nl\oldnl}}
\allowdisplaybreaks

\begin{document}

\author{\IEEEauthorblockN{Hamid Mirzaei}
\IEEEauthorblockA{Dept. of Computer Science\\
University of California, Irvine\\
mirzaeib@uci.edu}
\and
\IEEEauthorblockN{Tony Givargis}
\IEEEauthorblockA{Dept. of Computer Science\\
University of California, Irvine\\
givargis@uci.edu}
}

\title{Fine-grained acceleration control for autonomous intersection management using deep reinforcement learning}

\maketitle
\begin{abstract}
Recent advances in combining deep learning and Reinforcement Learning have shown a promising path for designing new control agents that can learn optimal policies for challenging control tasks. These new methods address the main limitations of conventional Reinforcement Learning methods such as customized feature engineering and small action/state space dimension requirements. In this paper, we leverage one of the state-of-the-art Reinforcement Learning methods, known as Trust Region Policy Optimization, to tackle intersection management for autonomous vehicles. We show that using this method, we can perform fine-grained acceleration control of autonomous vehicles in a grid street plan to achieve a global design objective.
\end{abstract}

\section{Introduction}
Previous works on autonomous intersection management (AIM) in urban areas have mostly focused on intersection arbitration as a shared resource among a large number of autonomous vehicles. In these works \cite{hausknecht2011autonomous}\cite{dresner2008multiagent}, high-level control of the vehicles is implemented such that the vehicles are self-contained agents that only communicate with the intersection management agent to reserve space-time slots in the intersection. This means that low-level vehicle navigation which involves acceleration and speed control is performed by each individual vehicle independent of other vehicles and intersection agents. This approach is appropriate for minor arterial roads where a large number of vehicles utilize the main roads at similar speeds while the adjacent intersections are far away.

In scenarios involving local roads, where the majority of the intersections are managed by stop signs, the flow of traffic is more efficiently  managed using a fine-grained vehicle control methodology. For example, when two vehicles are crossing the intersection of two different roads at the same time, one vehicle can decelerate slightly to avoid collision with the other one or it can take another path to avoid confronting the other vehicle completely. Therefore, the nature of the AIM problem is a combination of route planning and real-time acceleration control of the vehicles. In this paper, we propose a novel AIM formulation which is the combination of route planning and fine-grained acceleration control. The main objective of the control task is to minimize travel time of the vehicles while avoiding collisions between them and other obstacles. In this context, since the movement of a vehicle is dependent on the other vehicles in the same vicinity, the motion data of all vehicles is needed in order to solve the AIM problem.

To explain the proposed AIM scheme, let us define a ``zone'' as a rectangular area consisting of a number of intersections and segments of local roads. An agent for each zone collects the motion data and generates the acceleration commands for all autonomous vehicles within the zone's boundary. All the data collection and control command generation should be done in real-time. This centralized approach cannot be scaled to a whole city, regardless of the algorithm used, due to the large number of vehicles moving in a city which requires enormous computational load and leads to other infeasible requirements such as low-latency communication infrastructure. Fortunately, the spatial independence (i.e., the fact that navigation of the vehicles in one zone is independent of the vehicles in another zone that is far enough away) makes AIM an inherently local problem. Therefore, we can assign an agent for each local zone in a cellular scheme.

The cellular solution nevertheless leads to other difficulties that should be considered for a successful design of the AIM system. One issue is the dynamic nature of the transportation problem. Vehicles can enter or leave a zone controlled by an agent or they might change their planned destinations from time to time. To cope with these issues, the receding horizon control method can be employed where the agent repeatedly recalculates the acceleration command over a moving time horizon to take into account the mentioned changes. Additionally, two vehicles that are moving toward the same point on the boundary of two adjacent zones simultaneously might collide because the presence of each vehicle is not considered by the agent of the adjacent zone. This problem can be solved by adequate overlap between adjacent zones. Furthermore, any  planned trip for a vehicle typically crosses multiple zones. Hence, a higher level planning problem should be solved first that determines the entry and exit locations of a vehicle in a zone.

In this paper we focus on the subproblem of acceleration control of the vehicles moving in a zone to minimize the total travel time. We use a deep reinforcement learning (RL) approach to tackle the fine-grained acceleration control problem since conventional control methods are not applicable because of the non-convex collision avoidance constraints \cite{frese2011comparison}. Furthermore, if we want to incorporate more elements into the problem, such as obstacles or reward/penalty terms for gas usage, passenger comfort, etc., the explicit modeling becomes intractable and an optimal control law derivation will be computationally unattainable.

RL methods can address the above mentioned limitations caused by the explicit modeling requirement and conventional control method limitations. The main advantage of RL is that most of the RL algorithms are ``model-free'' or at most need a simulation model of the physical system which is easier to develop than an explicit model. Moreover, the agent can learn optimal policies just by interacting with the environment or executing the simulation model. However, conventional RL techniques are only applicable in small-scale problem settings and require careful design of approximation functions. Emerging Deep RL methods \cite{duan2016benchmarking} that leverage the deep neural networks to automatically extract features seem like promising solutions to shortcomings of the classical RL methods.  

The main contributions of this paper are: (1) Definition and formulation of the AIM problem for local road settings where vehicles arecoordinated by fine-grained acceleration commands. (2) Employing TRPO proposed in \cite{schulman2015trust} to solve the formulated AIM problem. (3) Incorporating collision avoidance constraint in the definition of RL environment as a safety mechanism.

\section{Related Work}

Advances in autonomous vehicles in recent years have revealed a portrait of a near future in which all vehicles will be driven by artificially intelligent agents. This emerging technology calls for an intelligent transportation system by redesigning the current transportation system which is intended to be used by human drivers. One of the interesting topics that arises in intelligent transportation systems is AIM. Dresner et al. have proposed a multi-agent AIM system in which vehicles communicate with intersection management agents to reserve a dedicated spatio-temporal trajectory at the intersection \cite{dresner2008multiagent}.

In \cite{mladenovic2013self}, authors have proposed a self-organizing control framework in which a cooperative multi-agent control scheme is employed in addition to each vehicle's autonomy. The authors have proposed a priority-level system to determine the right-of-way through intersections based on vehicles' characteristics or intersection constraints.

Zohdy et al. presented an approach in which the Cooperative Adaptive Cruise Control (CACC) systems are leveraged to minimize delays and prevent clashes \cite{zohdy2012intersection}. In this approach, the intersection controller communicates with the vehicles to recommend the optimal speed profile based on the vehicle's characteristics, motion data, weather conditions and intersection properties. Additionally, an optimization problem is solved to minimize the total difference of actual arrival times at the Intersection and the optimum times subject to conflict-free temporal constraints.

A decentralized optimal control formulation is proposed in \cite{malikopoulos2016decentralized} in which the acceleration/deceleration of the vehicles are minimized subject to collision avoidance constraints. 

Makarem et al. introduced the notion of fluent coordination where smoother trajectories of the vehicles are achieved through a navigation function to coordinate the autonomous vehicles along predefined paths with expected arrival time at intersections to avoid collisions.

In all the aforementioned works, the AIM problem is formulated for only one intersection and no global minimum travel time objective is considered directly. Hausknecht et al. extended the approach proposed in \cite{dresner2008multiagent} to multi-intersection settings via dynamic traffic assignment and dynamic lane reversal \cite{hausknecht2011autonomous}. Their problem formulation is based on intersection arbitration which is well suited to main roads with a heavy load of traffic.

In this paper, for the first time, we introduce fine-grained acceleration control for AIM. In contrast to previous works, Our proposed AIM scheme is applicable to local road intersections. We also propose an RL-based solution using Trust Region Policy Optimization to tackle the defined AIM problem.

\section{Reinforcement Learning}\label{sec:rl}

\begin{figure}[t]
  \centering
  \includegraphics[width= 0.56 \mycolumnwidth]{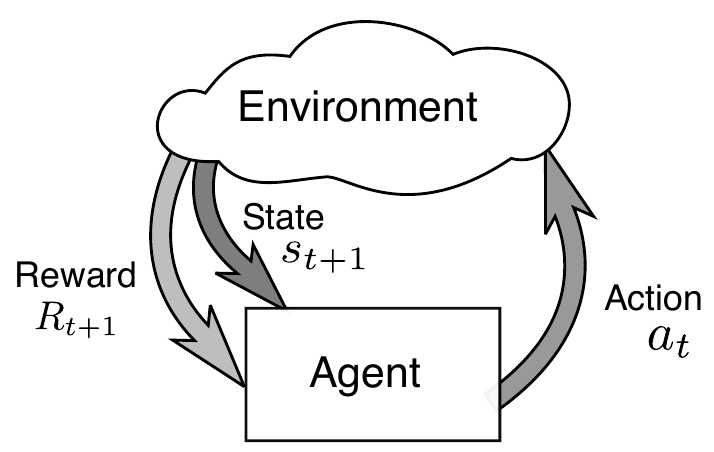}
  \caption{Agent-Environment interaction model in RL}
  \label{fig:rl}
\end{figure} 

In this section, we briefly review RL and introduce the notations used in the rest of the paper. In Fig. \ref{fig:rl}, the agent-environment model of RL is shown. The ``agent'' interacts with the ``environment'' by applying ``actions'' that influence the environment state at the future time steps and observes the state and ``reward'' in the next time step resulting from the action taken. The ``return'' is defined as the sum of all the rewards from the current step to the end of current ``episode'':

\begin{align}\label{eq:det_ret}
  G_t = \sum_{i=t}^T r_i
\end{align}
where $r_i$ are future rewards and $T$ is the total number of steps in the episode. An ``episode'' is defined as a sequence of agent-environment interactions. In the last step of an episode the control task is ``finished.'' Episode termination is defined specifically for the control task of the application. 

For example, in the cart-pole balancing task, the agent is the controller, the environment is the cart-pole physical system, the action is the force command applied to the cart, and the reward can be defined as $r=1$ as long as the pole is nearly in an upright position and a large negative number when the pole falls. 
The system states are cart position, cart speed, pole angle and pole angular speed. 
The agent task is to maximize the return $G_t$, which is equivalent to prevent pole from falling for the longest possible time duration.

In RL, a control policy is defined as a mapping of the system state space to the actions:
\begin{align}\label{eq:det_pol}
  a &= \pi(s) 
\end{align}
where $a$ is the action, $s$ is the state and  $\pi$ is the policy. An optimal policy is one that maximizes the return for all the states, i.e.:
\begin{align}
  v_{\pi^*}(s) & \ge v_\pi (s),\quad \text{for all}\,\, s,\pi \label{eq:opt_pol}
\end{align}
where $v$ is the return function defined as the return achievable from state $s$ by following policy $\pi$. Equation (\ref{eq:opt_pol}) means that the expected return under optimal policy $\pi^*$ is equal to or greater than any other policy for all the system states. 

The concepts mentioned above to introduce RL are all applicable to deterministic cases, but generally we should be able to deal with inherent system uncertainty, measurement noise, or both. Therefore, we model the system as a Markov Decision Process (MDP) assuming that the environment has the \textit{Markov property} \cite{sutton1998reinforcement}. However, contrary to most of the control design methods, many RL algorithms do not require the models to be known beforehand. The elimination of the requirement to model the system under control is a major strength of RL. 

A system has the Markov property if at a certain time instant, $t$, the system history can be captured in a set of \textit{state variables}. Therefore, the next state of the system has a distribution which is only conditioned on the current state and the taken action at the current time, i.e.:
\begin{align}
  s_{t+1} \sim P(s_{t+1}|s_t,a_t)
\end{align}
The Markov property holds for many cyber-physical system application domains and therefore MDP and RL can be applied as the control algorithm. We can also define the stochastic policy which is a generalized version of (\ref{eq:det_pol}) as a probability distribution of actions conditioned on the current state, i.e.:
\begin{align}
  a_{t} \sim \pi(a_{t}|s_t)  
\end{align}
The expected return function which is the expected value of `return' defined in (\ref{eq:det_ret}) can be written as:
\begin{align}
  v_{\pi}(s_t) =  \E{a_{\tau} \sim \pi, \tau \ge t}{\sum_{i=0}^\infty \gamma^i r_{t+i}} 
\end{align}
This equation is defined for infinite episodes and the constant $0<\gamma<1$ is introduced to ensure that the defined expected return is always a finite value, assuming the returns are bounded.

Another important concept in RL is the action-value function, $Q_\pi(s,a)$ defined as the expected return (value) if action $a_t$ is taken at time $t$ under policy $\pi$:
\begin{align}
  Q_{\pi}(s_t, a_t) =  \E{a_{\tau} \sim \pi, \tau > t}{\sum_{i=0}^\infty \gamma^i r_{t+i}} 
\end{align}

There are two main categories of methods to find the optimal policy. In the first category, $Q_\pi(s,a)$ is parameterized as $Q^\theta_\pi(s,a)$ and the optimal action-value parameter vector $\theta$ is estimated in an iterative process. The optimal policy can be defined implicitly from $Q_\pi(s,a)$. For example, the greedy policy is the one that maximizes $Q_\pi(s,a)$ in each step:
\begin{align}
  a_t = \argmax_a\left\{Q_\pi(s,a)\right\}
\end{align}
In the second category, which is called policy optimization and has been successfully applied to large-scale and continuous control systems \cite{duan2016benchmarking}, the policy is parameterized directly as $\pi^\theta(a_t|s_t)$ and the parameter vector of the optimal policy $\theta$ is estimated. The Trust Region Policy Method (TRPO) \cite{schulman2015trust} is an example of the second category of methods that guarantees monotonic policy improvement and is designed to be scalable to large-scale settings. In each iteration of TRPO, a number of MDP trajectories are simulated (or actually experienced by the agent) and $\theta$ is updated to improve the policy. A high level description of TRPO is shown in algorithm \ref{alg:trpo}.

\begin{algorithm}[!b]
 \KwData{$\mathscr{S}$ \Comment {Actual system or Simulation model} \\
   $\pi^\theta$ \Comment {Parameterized Policy}}
 \KwResult{$\theta^*$ \Comment {Optimal parameters}}
 \Repeat{no more improvements}{
   Use $\mathscr{S}$ to generate trajectories of the system using current $\pi^\theta$\;
   Perform one iteration of policy optimization using Monte Carlo method to get $\theta_{new}$ \;
   $\theta \gets \theta_{new}$
 }
 \Return{$\theta$}
 \caption{\footnotesize High-Level description of Trust Region Optimization}
 \label{alg:trpo}
\end{algorithm}

\begin{figure}[t]
  \centering
  \includegraphics[width= 0.7 \mycolumnwidth]{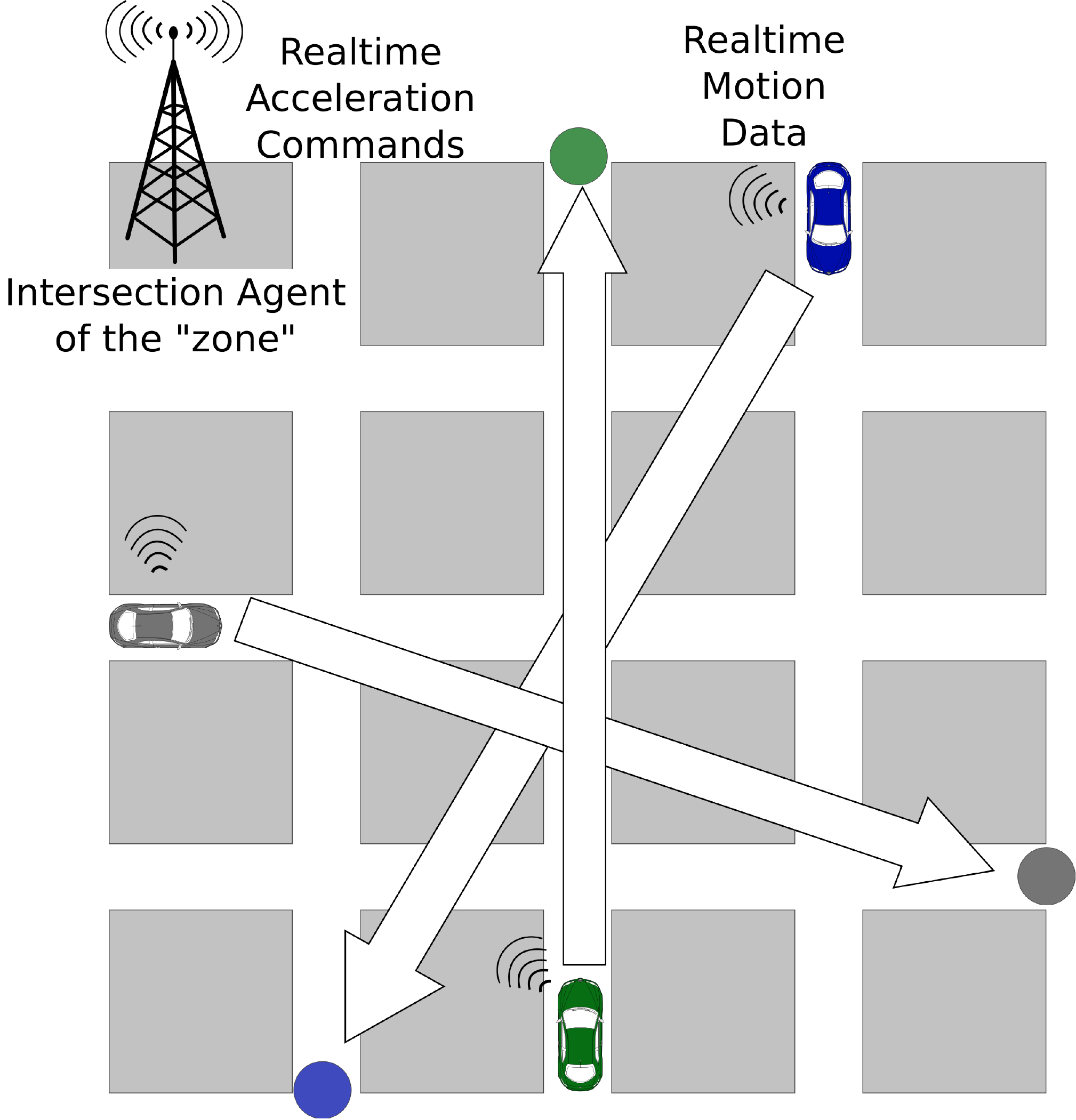}
  \caption{Intersection Management Problem. The goal of the problem is to navigate the vehicles from the sources to destinations in minimum time with no collisions.}
  \label{fig:prob}
\end{figure}

\section{Problem Statement} \label{sec:problem-statement}

There is a set of vehicles in a grid street plan area consisting of a certain number of intersections. For simplicity, we assume that all the initial vehicle positions and the desired destinations are located at the intersections. There is a control agent for the entire area. The agent's task is to calculate the acceleration command for the vehicles in real-time (see Fig. \ref{fig:prob}). We assume that there are no still or moving obstacles other than vehicles' or street boundaries.

The input to the agent is the real-time state of the vehicles which consists of their positions and speeds. We are assuming that vehicles are point masses and their angular dynamics are ignored. However, to take the collision avoidance in the problem formulation, we define a safe radius for each vehicle and no objects (vehicles or street boundaries) should be closer than the safe radius to the vehicle.

The objective is to drive all the vehicles to their respective destinations in a way that the total travel time is minimized. Furthermore, no collision should occur between any two vehicles or a vehicle and the street boundaries.

To minimize the total travel time, a positive reward is assigned to the terminal state in which all the vehicles approximately reach the destinations within some tolerance. A discount factor $\gamma$ strictly less than one is used. Therefore, the agent should try to reach the terminal state as fast as possible to maximize the discounted return. However, by using only this reward, too many random walk trajectories are needed to discover the terminal state. Therefore, a negative reward is defined for each state, proportional to the total distance of the vehicles to their destinations as a hint of how far the terminal state is. This negative reward is not in contradiction with the main goal which is to minimize total travel time.

To avoid collisions, two different approaches can be considered: we can add large negative rewards for the collision states or we can incorporate a collision avoidance mechanism into the environment model. Our experiments show that the first approach makes the agent too conservative about moving the vehicles to minimize the probability of collisions. This might lead to extremely slow learning which makes it infeasible. Furthermore, collisions are inevitable even with large negative rewards which limits the effectiveness of learned policies in practice. 

For the above mentioned reasons, the second approach is employed, i.e. the safety mechanism that is used in practice is included in the environment definition. The safety mechanism is activated whenever two vehicles are too close to each other or a vehicle is too close to the street boundary. In these cases, the vehicle built-in collision avoidance system will control the vehicle's acceleration and the acceleration commands from the RL agent are ignored as long as the distance is near the allowed safe radius of the vehicle. In the agent learning process these cases are simulated in a way that the vehicles come to a full stop when they are closer than the safe radius to another vehicle or boundary. By applying this heuristic in the simulation model, the agent should avoid any ``near collision'' situations explained above because the deceleration and acceleration cycles take a lot of time and will decrease the expected return.

Based on the problem statement explained above, we can describe the RL formulation in the rest of the subsection. The state is defined as the following vector:
\begin{align}
  \vec{s_t} = \left( x^1_t, y^1_t,{v^1_x}_t, {v^1_y}_t, \dots,x^n_t, y^n_t,{v^n_x}_t, {v^n_y}_t \right)^\intercal \label{eq:st_vec}
\end{align}
where $(x^i_t, y^i_t)$ and $({v^i_x}_t, {v^i_y}_t)$ are the position and speed of vehicle $i$ at time $t$. The action vector is defined as:
\begin{align}
  \vec{a_t} = \left({a^1_x}_t, {a^1_y}_t, \dots,{a^n_x}_t, {a^n_y}_t \right)^\intercal \label{eq:ac_vec}
\end{align}
where $({a^i_x}_t, {a^i_y}_t)$ is the acceleration command of vehicle $i$ at time $t$. The reward function is defined as:
\begin{align}
  r(s) = 
\begin{cases}
    1 \quad \text{if} \quad \norm{(x^i-{d^i_x}, y^i-{d^i_x})^\intercal} < \eta \,\, (1 \le i \le n)   \\
    - \alpha \sum_{i=1}^n \norm{(x^i-{d^i_x}, y^i-{d^i_x})^\intercal} \quad \text{otherwise}  \\  
\end{cases}
\end{align}
where $({d^i_x}, {d^i_y})$ is the destination coordinates of vehicle $i$, $\eta$ is the distance tolerance and $\alpha$ is a positive constant.

Assuming no collision occurs, the state transition equations for the environment are defined as follows:
\begin{align}
  x^i_{t+1} &= \text{sat}_{\underline{x},\overline{x}} (x^i_t + h  {v_x}^i_t ) \nonumber \\
  y^i_{t+1} &= \text{sat}_{\underline{y},\overline{y}} (y^i_t + h  {v_y}^i_t ) \nonumber \\
  {v_x}^i_{t+1} &= \text{sat}_{\underline{{v_m}},\overline{{v_m}}} ({v_x}^i_t + h  {a_x}^i_t ) \nonumber \\
  {v_y}^i_{t+1} &= \text{sat}_{\underline{{v_m}},\overline{{v_m}}} ({v_y}^i_t + h  {a_y}^i_t ) \label{eq:state_trans}
\end{align}
where $h$ is the sampling time, $(\underline{x},\overline{x},\underline{y},\overline{y})$ defines area limits, $v_m$ is the maximum speed and $\text{sat}_{\underline{w},\overline{w}}(.)$ is the saturation function defined as:
\begin{align}
  \text{sat}_{\underline{w},\overline{w}}(x) = 
\begin{cases}
  \underline{w} & x \le \underline{w} \\
  \overline{w} & x \ge \overline{w} \\
  x & \text{otherwise}.
\end{cases}
\end{align}

To model the collisions, we should check certain conditions and set the speed to zero. A more detailed description of collision modeling is presented in Algorithm \ref{alg:trans}.

\begin{algorithm}[!b]
 \KwData{$s_t$ \Comment {State at time $t$} \\
   $a_t$ \Comment {Action at time $t$}}
 \KwResult{$s_{t+1}$ \Comment {State at time $t+1$}}
  ${a_x}^i_t \gets \text{sat}_{\underline{a_m},\overline{a_m}} ({a_x}^i_t)$ \;
  ${a_y}^i_t \gets \text{sat}_{\underline{a_m},\overline{a_m}} ({a_y}^i_t)$ \;
  $s_{t+1} \gets$ updated state using (\ref{eq:state_trans}) \;
  $v_{c1} \gets$ find all the vehicles colliding with street boundaries \;
  speed elements of $v_{c1}$in $s_{t+1} \gets 0$ \;
  location elements of $v_{c1}$in $s_{t+1} \gets$ closest point on the street boundary with the margin of $\epsilon$ \;
  $v_{c2} \gets$ find all the vehicles colliding with some other vehicle \;
  speed elements of $v_{c2}$in $s_{t+1} \gets 0$ \;
  location elements of $v_{c2}$in $s_{t+1} \gets$ pushed back location with the distance of $2 \times$ safe radius to the collided vehicle\;
  \Return{$s_{t+1}$
}
 \caption{\footnotesize State Transition Function}
 \label{alg:trans}
\end{algorithm}

\subsection{Solving the AIM problem using TRPO}

The simulation model can be implemented based on the RL formulation described in Section \ref{sec:problem-statement}. To use TRPO, we need a parameterized stochastic policy, $\pi^\theta(a_t|s_t)$,  in addition to the simulation model. The policy should specify the probability distribution for each element of the action vector defined in (\ref{eq:ac_vec}) as a function of the current state $s_t$.

We have used the sequential deep neural network (DNN) policy representation as described in \cite{schulman2015trust}. The input layer receives the state containing the position and speed of the vehicles (defined in (\ref{eq:st_vec})). There are a number of hidden layers, each followed by $\tanh$ activation functions \cite{karlik2011performance}. Finally, the output layer generates the mean of a gaussian distribution for each element of the action vector. 

To execute the optimal policy learned by TRPO in each sampling time, the agent calculates the forward-pass of DNN using the current state. Next, assuming that all the action elements have the same variance, the agent samples from the action gaussian distributions and applies the sampled actions to the environment as the vehicle acceleration commands.

\section{Evaluation}
\subsection{Baseline Method}
To the best of our knowledge there is no other solution proposed for the fine-grained acceleration AIM problem introduced in this paper. Therefore, we use conventional optimization methods to study how close the proposed solution is to the optimal solution. Furthermore, we will see that the conventional optimization is able to solve the AIM problem only for very small-sized problems. This confirms that the proposed RL-based solution is a promising alternative to the conventional methods.

Theoretically, the best solution to the problem defined in section \ref{sec:problem-statement} can be obtained if we reformulate it as a conventional optimization problem. The following equations and inequalities describe the AIM optimization problem:

\begin{align}
\vec{a_t}^* &= \argmax_{\vec{a_t}} \sum_{t=0}^{T-1} \sum_{i=1}^n \norm{(x^i_t-{d^i_x}, y^i_t-{d^i_x})^\intercal} \\
\text{s. t.} \quad 
\underline{x} &\le x^i_t \le \overline{x} \quad (1 \le i \le n) \label{eq:xlim} \\
\underline{y} &\le y^i_t \le \overline{y} \quad (1 \le i \le n)\\
\underline{v_m} &\le {v_x}^i_t \le \overline{v_m} \quad (1 \le i \le n)\\
\underline{v_m} &\le {v_y}^i_t \le \overline{v_m} \quad (1 \le i \le n)\\
\underline{a_m} &\le {a_x}^i_t \le \overline{a_m} \quad (1 \le i \le n)\\
\underline{a_m} &\le {a_y}^i_t \le \overline{a_m} \quad (1 \le i \le n) \label{eq:aylim} \\
-\floor*{N/2} &\le r^i_t \le \floor*{N/2} \quad (1 \le i \le n, r^i_t \in \mathbb{Z})\\ 
-\floor*{M/2} &\le c^i_t \le \floor*{M/2} \quad (1 \le i \le n, c^i_t \in \mathbb{Z})\\ 
x^i_0 &= s^i_x,\, y^i_0 = s^i_y \quad (1 \le i \le n)  \label{eq:init_pos} \\
x^i_{T-1} &= d^i_x,\, y^i_{T-1} = d^i_y \quad (1 \le i \le n)  \label{eq:finals} \\
{v_x}^i_0 &= 0,\, {v_y}^i_0 = 0 \quad (1 \le i \le n)  \label{eq:init_vel} \\
x^i_{t+1} &= x^i_{t} + {v_x}^i_t . h \quad (1 \le i \le n)  \label{eq:posx_dyn} \\
y^i_{t+1} &= y^i_{t} + {v_y}^i_t . h \quad (1 \le i \le n) \\ 
{v_x}^i_{t+1} &= {v_x}^i_{t} + {a_x}^i_t . h \quad (1 \le i \le n) \\
{v_y}^i_{t+1} &= {v_y}^i_{t} + {a_y}^i_t . h \quad (1 \le i \le n)  \label{eq:vely_dyn} \\
(x^i_t &-x^j_t)^2+(y^i_t-y^j_t)^2 \ge (2R)^2 \quad (1 \le i < j \le n)  \label{eq:cartocar} \\
|x^i_t &- c^i_t . b_w| \le (\frac{l}{2}-R) \,\, \text{or} \nonumber \\
|y^i_t &- r^i_t . b_h| \le (\frac{l}{2}-R) \,\, (1 \le i \le n) \label{eq:cartoblock}
\end{align}
where $r^i_t$ and $c^i_t$ are the row number and column number of vehicle at time $t$, respectively, assuming the zone is a perfect rectangular grid; N and M are the number of rows and columns, respectively; $b_w$ and $b_h$ are block width and block height; $l$ is the street width; $R$ is the vehicle clearance radius; $T$ is number of sampling times; and $({s^i_x}, {s^i_y})$ is the source coordinates of vehicle $i$.

\begin{figure}[t]
  \centering
\begin{subfigure}{.30\textwidth}
  \centering
  \includegraphics[width= 0.25 \mycolumnwidth]{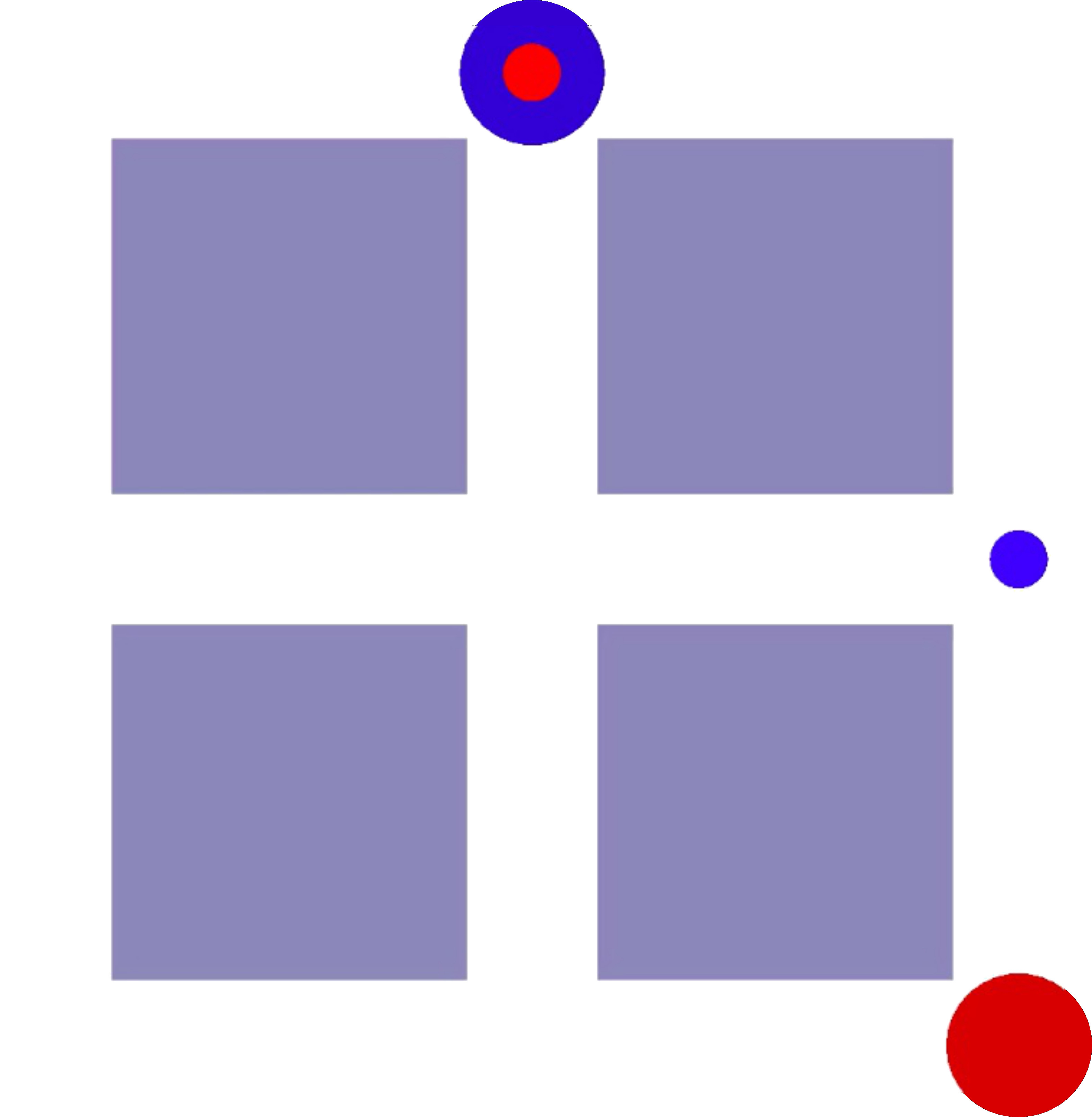}
  \caption{}
\end{subfigure}
\begin{subfigure}{.5\textwidth}
  \centering
  \includegraphics[width= 0.5 \mycolumnwidth]{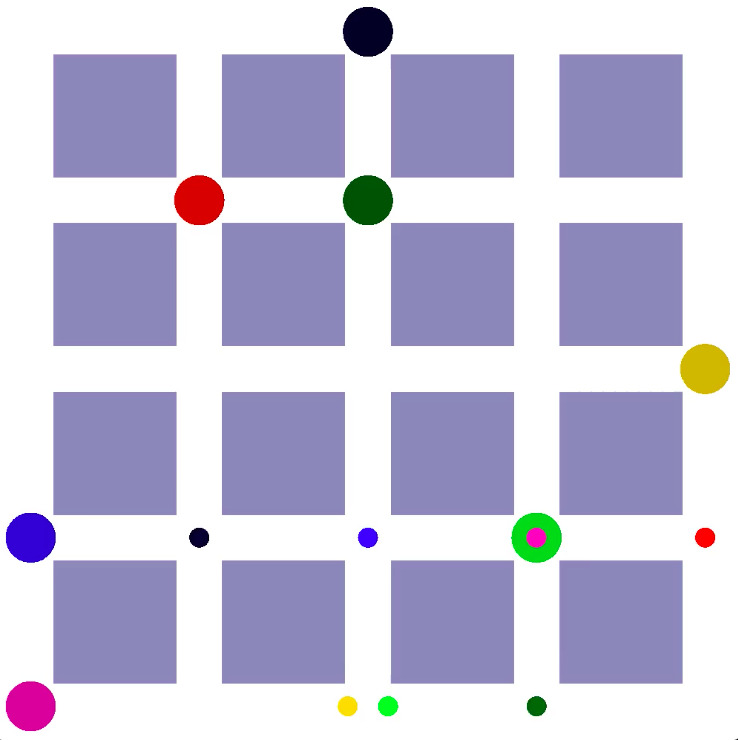}
  \caption{}
\end{subfigure}
  \caption{Initial setup of each episode. Small circles are the sources and big circles are the destinations. (a) small example (b) large example} 
  \label{fig:init}
\end{figure} 

In the above mentioned problem setting, (\ref{eq:xlim}) to (\ref{eq:aylim}) are the physical limit constraints. (\ref{eq:init_pos}) to (\ref{eq:init_vel}) describe the initial and final conditions. (\ref{eq:posx_dyn}) to (\ref{eq:vely_dyn}) are dynamic constraints. (\ref{eq:cartocar}) is the vehicle-to-vehicle collision avoidance constraint and finally (\ref{eq:cartoblock}) is the vehicle-to-boundaries collision avoidance constraint.

\begin{figure*}[!t]
  \centering
  \includegraphics[width=0.4 \textwidth]{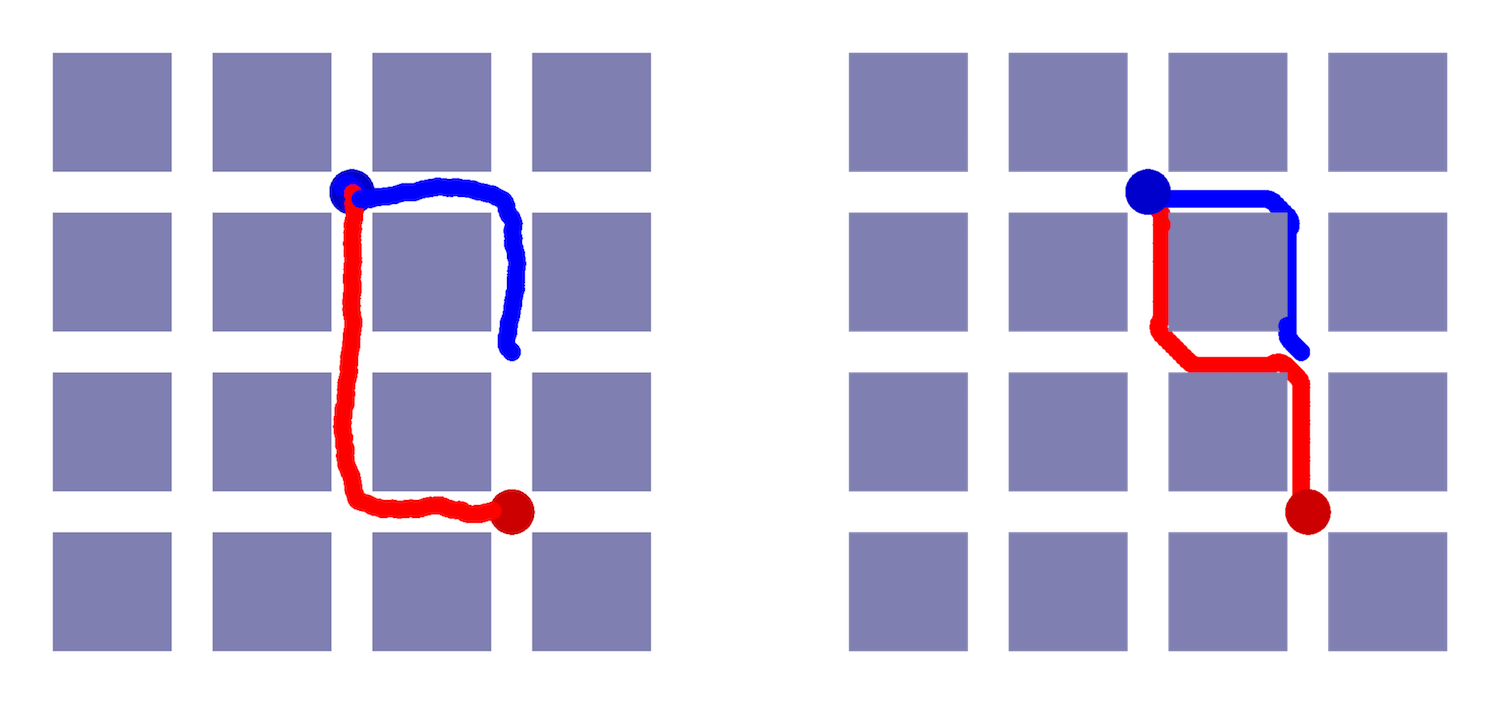}
  \caption{Learnt policy by (left)RL agent and (right)the baseline method for the small example}
  \label{fig:learning_comp}
\end{figure*} 
\begin{figure*}[!t]
  \centering
  \includegraphics[width=0.8 \textwidth]{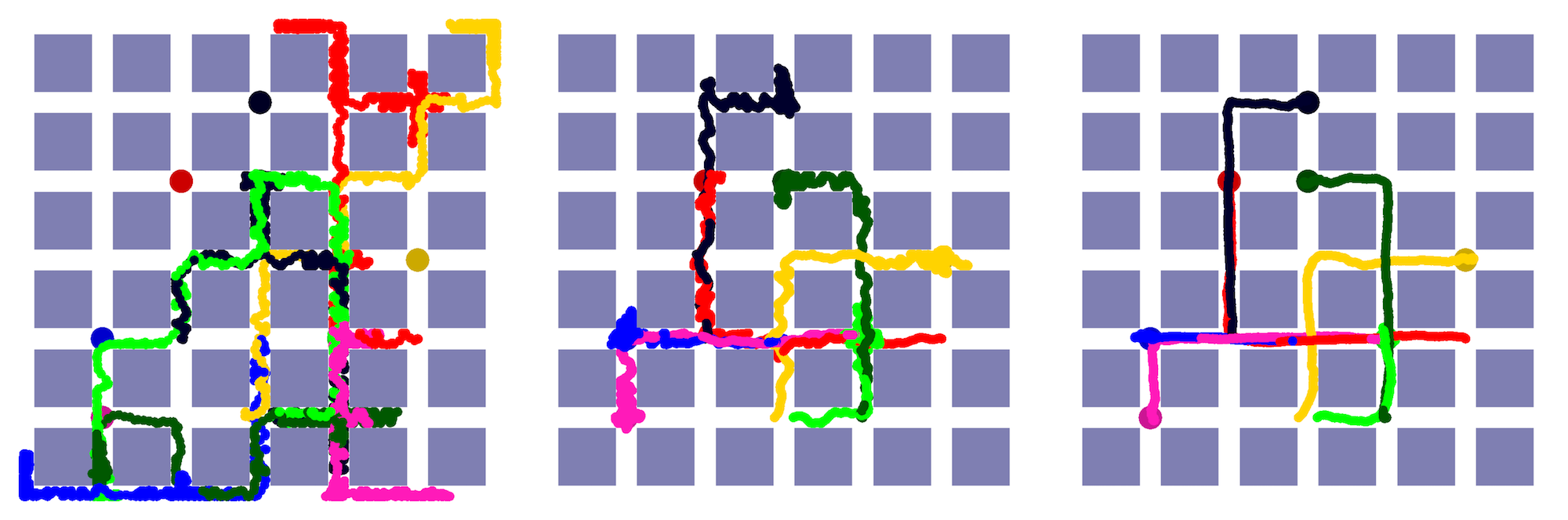}
  \caption{Learnt policy by the AIM agent at different iterations of the training. left: begging of the training, middle: just after the fast learning phase in Fig. \ref{fig:return}. right: end of the training.}
  \label{fig:learning}
\end{figure*} 

The basic problem with the above formulation is that constraint (\ref{eq:cartocar}) leads to a non-convex function and convex optimization algorithms cannot solve this problem. Therefore, a Mixed-Integer Nonlinear Programming (MINLP) algorithm should be used to solve this problem. Our experiments show that even a small-sized problem with two vehicles and 2$\times$2 grid cannot be solved with an MINLP algorithm, i.e. AOA\cite{hunting2011aimms}. To overcome this issue, we should reformulate the optimization problem using 1-norm and introduce new integer variables for the distance between vehicles using the ideas proposed in \cite{schouwenaars2001mixed}. 

To achieve the best convergence and execution time by using a Mixed-integer Quadratic Programming (MIQP), the cost function and all constraints should be linear or quadratic. Furthermore, the ``or'' logic in (\ref{eq:cartoblock}) should be implemented using integer variables. The full MIQP problem can be written as the following equations and inequalities:

\begin{align}
\vec{a_t}^* &= \argmax_{\vec{a_t}} \sum_{t=0}^{T-1} \sum_{i=1}^n (x^i_t-{d^i_x})^2 + (y^i_t-{d^i_x})^2 \label{eq:obj}\\
\text{s. t.} \quad 
&(\ref{eq:xlim}) \,\, \text{to} \,\, (\ref{eq:vely_dyn}) \nonumber \\
{b_x}^i_t, {b_y}^i_t &\in \{0, 1\}   \quad (1 \le i \le n) \\
{b_x}^i_t + {b_y}^i_t  &\ge 1 \quad (1 \le i \le n) \label{eq:orlogic2}\\
{c_x}^{i,j}_t, &{c_y}^{i,j}_t, {d_x}^{i,j}_t, {d_y}^{i,j}_t  \in \{0, 1\} \quad (1 \le i < j \le n) \label{eq:orlogic1}\\
{c_x}^{i,j}_t + &{c_y}^{i,j}_t+ {d_x}^{i,j}_t+ {d_y}^{i,j}_t  \ge 1 \quad (1 \le i < j \le n) \\
x^i_t - x^j_t &\ge 2R {c_x}^{i,j}_t - \mathcal{M} (1-{c_x}^{i,j}_t) \quad (1 \le i < j \le n) \label{eq:car2car_s}\\
x^i_t - x^j_t &\le -2R {d_x}^{i,j}_t + \mathcal{M} (1-{d_x}^{i,j}_t) \quad (1 \le i < j \le n) \\
y^i_t - y^j_t &\ge 2R {c_y}^{i,j}_t - \mathcal{M} (1-{c_y}^{i,j}_t) \quad (1 \le i < j \le n) \\
y^i_t - y^j_t &\le -2R {d_y}^{i,j}_t + \mathcal{M} (1-{d_y}^{i,j}_t) \quad (1 \le i < j \le n) \label{eq:car2car_e}\\
x^i_t - c^i_t b_w &\le (\frac{l}{2} - R){b_x}^i_t + \mathcal{M} (1-{b_x}^i_t)  \quad (1 \le i \le n) \label{eq:car2block_s}\\
x^i_t - c^i_t b_w &\ge -(\frac{l}{2} - R){b_x}^i_t - \mathcal{M} (1-{b_x}^i_t)  \quad (1 \le i \le n) \\
y^i_t - c^i_t b_w &\le (\frac{l}{2} - R){b_y}^i_t + \mathcal{M} (1-{b_y}^i_t)  \quad (1 \le i \le n) \\
y^i_t - c^i_t b_w &\ge -(\frac{l}{2} - R){b_y}^i_t - \mathcal{M} (1-{b_y}^i_t)  \quad (1 \le i \le n) \label{eq:car2block_e}
\end{align}
where $\mathcal{M}$ is a large positive number. 

(\ref{eq:car2car_s}) to (\ref{eq:car2car_e}) represent the vehicle-to-vehicle collision avoidance constraint using 1-norm:
\begin{align}
  \norm{(x^i_t, y^i_t)^\intercal-(x^j_t, y^j_t)^\intercal}_1 \ge 2R
\end{align}
for any two distinct vehicles $i$ and $j$. This constraint is equivalent to the following:
\begin{align}
  |x^i_t-x^j_t| \ge 2R \,\, \text{or} \,\,   |y^i_t-y^j_t| \ge 2R \quad \forall t,(1 \le i < j \le n) \label{eq:car2car_abs}
\end{align}
The absolute value function displayed in (\ref{eq:car2car_abs}) should be replaced by logical ``or'' of two linear conditions to avoid nonlinearity. Therefore we have the following four constraints represented by (\ref{eq:car2car_s}) to (\ref{eq:car2car_e}):
\begin{align}
&x^i_t-x^j_t \ge 2R \,\, \text{or} \,\, x^i_t-x^j_t \le -2R \,\, \text{or}  \nonumber \\
&y^i_t-y^j_t \ge 2R \,\, \text{or} \,\, y^i_t-y^j_t \le -2R  \quad \forall t,(1 \le i < j \le n) \label{eq:car2car_mi}
\end{align}
(\ref{eq:orlogic1}) implements the ``or'' logic required in (\ref{eq:car2car_mi}). 

(\ref{eq:car2block_s}) to (\ref{eq:car2block_e}) describe the vehicle-to-boundaries collision avoidance constraint:
\begin{align}
  |x^i_t - c^i_t b_w| &\le (\frac{l}{2} - R){b_x}^i_t \,\, \text{or} \nonumber \\
  |y^i_t - r^i_t b_w| &\le (\frac{l}{2} - R){b_y}^i_t \quad \forall t,(1 \le i \le n)
\end{align}
which is equivalent to:
\begin{align}
  ( x^i_t - c^i_t b_w &\le (\frac{l}{2} - R) {b_x}^i_t \,\, \text{and} \,\, x^i_t - c^i_t b_w \ge -(\frac{l}{2} - R) {b_x}^i_t ) \,\, \text{or} \nonumber \\
  ( y^i_t - r^i_t b_w &\le (\frac{l}{2} - R) {b_y}^i_t \,\, \text{and} \,\, y^i_t - r^i_t b_w \ge -(\frac{l}{2} - R) {b_y}^i_t ) \nonumber \\
\quad &\forall t,(1 \le i \le n) \label{eq:car2bound_abs}
\end{align}
The ``or'' logic in this constraint is realized in (\ref{eq:orlogic2}).

We will show in the next subsection that the explained conventional optimization formulation is not feasible except for very small-sized problems. Another limitation that makes the conventional method impractical is that this formulation works only for a perfect rectangular grid. However, the proposed RL method in this paper can be extended to arbitrary street layouts. 

\subsection{Simulation results}

\begin{table}[b!]
\centering
\caption{Parameter value settings for the experiment}
\label{tab:params}
\begin{tabular}{@{}ll@{}}
  \toprule
  Parameter & Value \\ \midrule
  discount factor($\gamma$) &  0.999 \\
  distance reward penalty factor($\alpha$) & 0.1 \\
  distance tolerance($\eta$)  & 0.05 \\
  maximum speed($v_m$) & 0.8 \\
  maximum acceleration($a_m$) & 30 \\
  sampling time($h$) & 10 (ms) \\
  maximum episode length & 200 \\
  vehicle safe radius & 0.02 
\end{tabular}
\end{table}

The implementation of the TRPO in rllab library \cite{duan2016benchmarking} is used to simulate the RL formulation of the AIM problem described in Section \ref{sec:problem-statement}. For this purpose, the AIM state transition and reward calculation are implemented as an OpenAI Gym \cite{r_openai} environment.

\begin{figure}[t]
  \centering
  \includegraphics[width= 0.85 \mycolumnwidth]{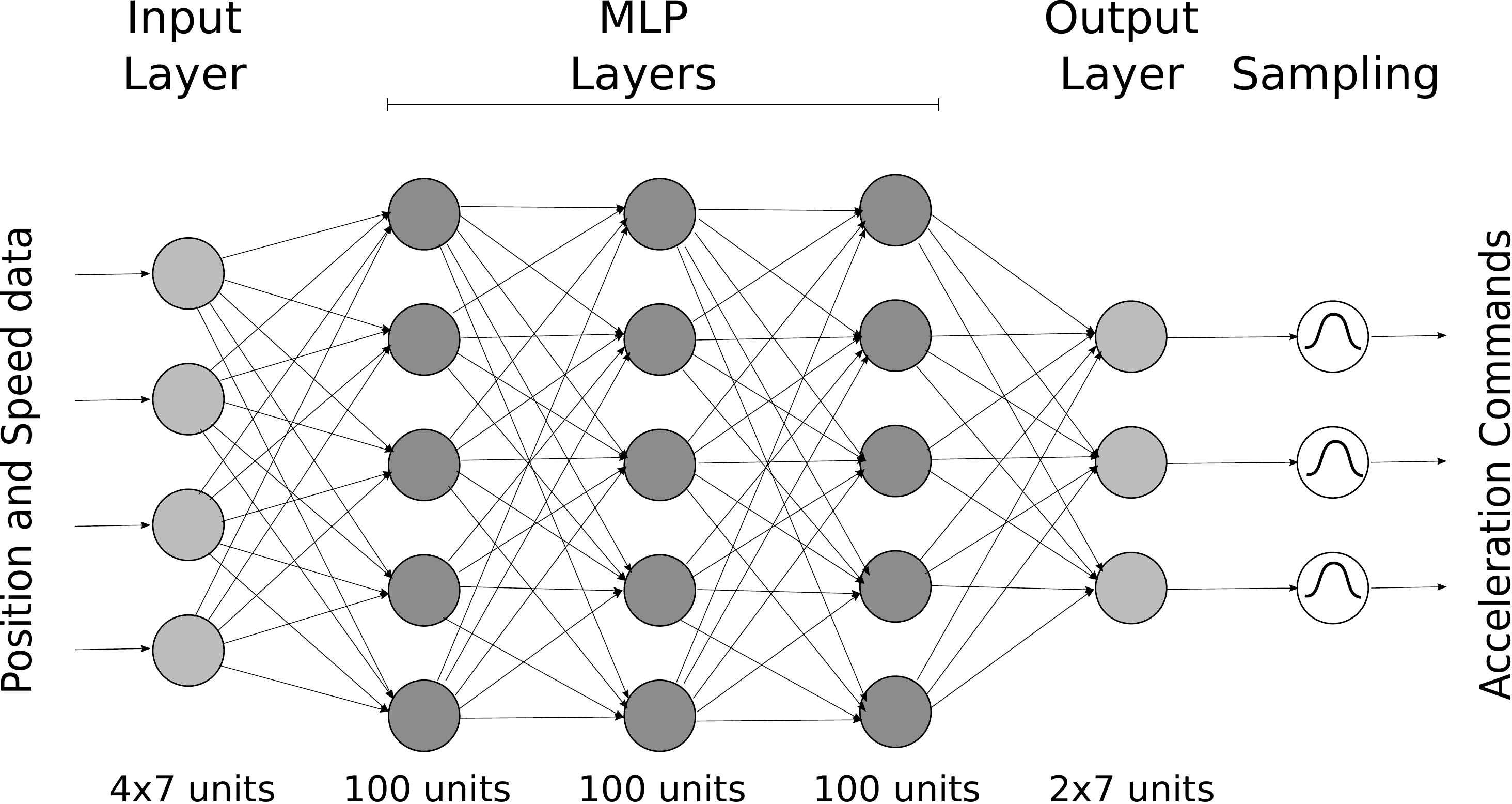}
  \caption{Neural network used in the simulations.}
  \label{fig:nn}
\end{figure}

The neural network used to approximate the policy is an MLP which consists of three hidden layers. Each hidden layer has 100 nodes (Fig. \ref{fig:nn}). Table \ref{tab:params} lists the parameters for the simulation. To speed up simulation, normalized units are used for the physical properties of the environment instead of real-world quantities.

\begin{table}[!t]
\centering
\caption{Total travel time obtained by baseline method and proposed method}
\label{my-label}
\begin{tabular}{l|l|l|}
\cline{2-3}
                                    & Baseline Method & Proposed Method \\ \hline
\multicolumn{1}{|l|}{Small example} & 1.79 (s)         & 2.43 (s)        \\ \hline
\multicolumn{1}{|l|}{Large Example} & no solution      & 11.2 (s)        \\ \hline
\end{tabular}
\label{tab:results}
\end{table}

Fig. \ref{fig:init} shows the small and large grid plans used for the simulation. The small and large circles represent the source and destination locations, respectively. The vehicles are placed at the intersections randomly at the beginning of each episode. The destinations are also chosen randomly. When the simulator is reset, the same set of source and destination locations are used.

The small grid can be solved both by the baseline method and by the proposed RL method. However, the large grid can only be solved by the RL method because the MIQP algorithm could not find a feasible solution (which is not optimal necessarily) and was stopped after around 68 hours and using 21 GB of system memory. On the other hand, the RL method can solve the problem using 560 MB of system memory and 101 MB of GPU memory. 

Table \ref{tab:results} and Fig. \ref{fig:learning_comp} show the comparison of proposed RL and baseline method results. In Table \ref{tab:results} the total travel time is provided for both methods and Fig. \ref{fig:learning_comp} shows the vehicles' trajectories by running the navigation policy obtained by both solutions for the small examples.

\begin{figure}[!t]
  \centering
  \includegraphics[width= 0.75 \mycolumnwidth]{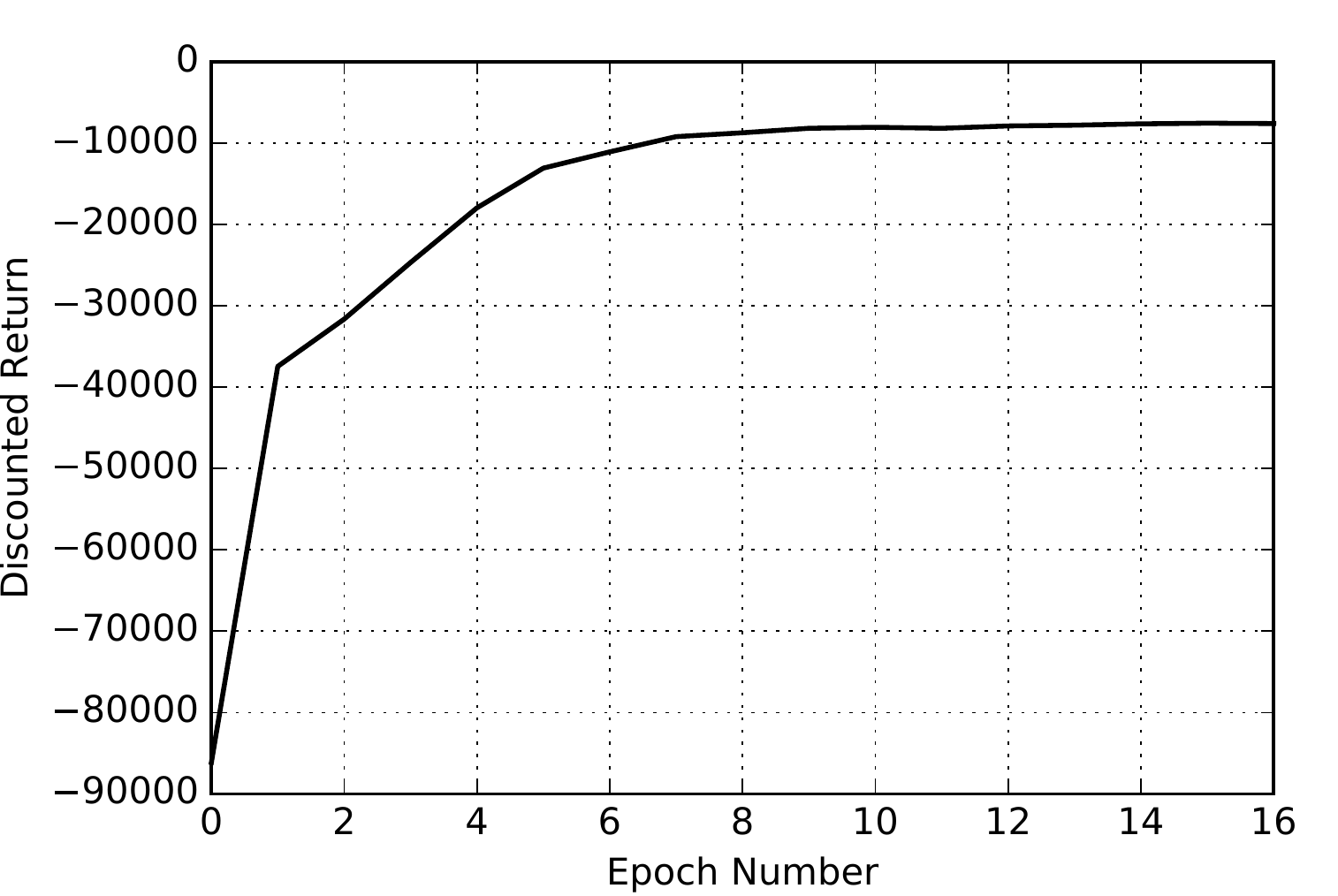}
  \caption{Learning curve of the AIM agent for large grid example. The discounted return is always a negative value because the return is the accumulated negative distance rewards and there is only one positive reward in the terminal state in which all the vehicles are at the destinations.}
  \label{fig:return}
\end{figure}

The learning curve of the RL agent which is the expected return vs the training epoch number is shown in Fig. \ref{fig:return} for the large grid example. This figure shows that the learning rate is higher at the beginning which corresponds to the stage where in the agent is learning the very basics of driving and avoiding collisions, but improving the policy towards the optimal policy takes considerably more time. The increase in learning occurs after two epochs when the agent discovers the policy that successfully drives all the vehicles to the destination and the positive terminal reward is gained. Moreover, the trajectories of vehicles are depicted in Fig. \ref{fig:learning} at three stages of the learning process, i.e. at the early stage, at epoch 2 where the learning curve slope suddenly decreases, and the end of the training.

\begin{figure}[!t]
  \centering
  \includegraphics[width= 0.75 \mycolumnwidth]{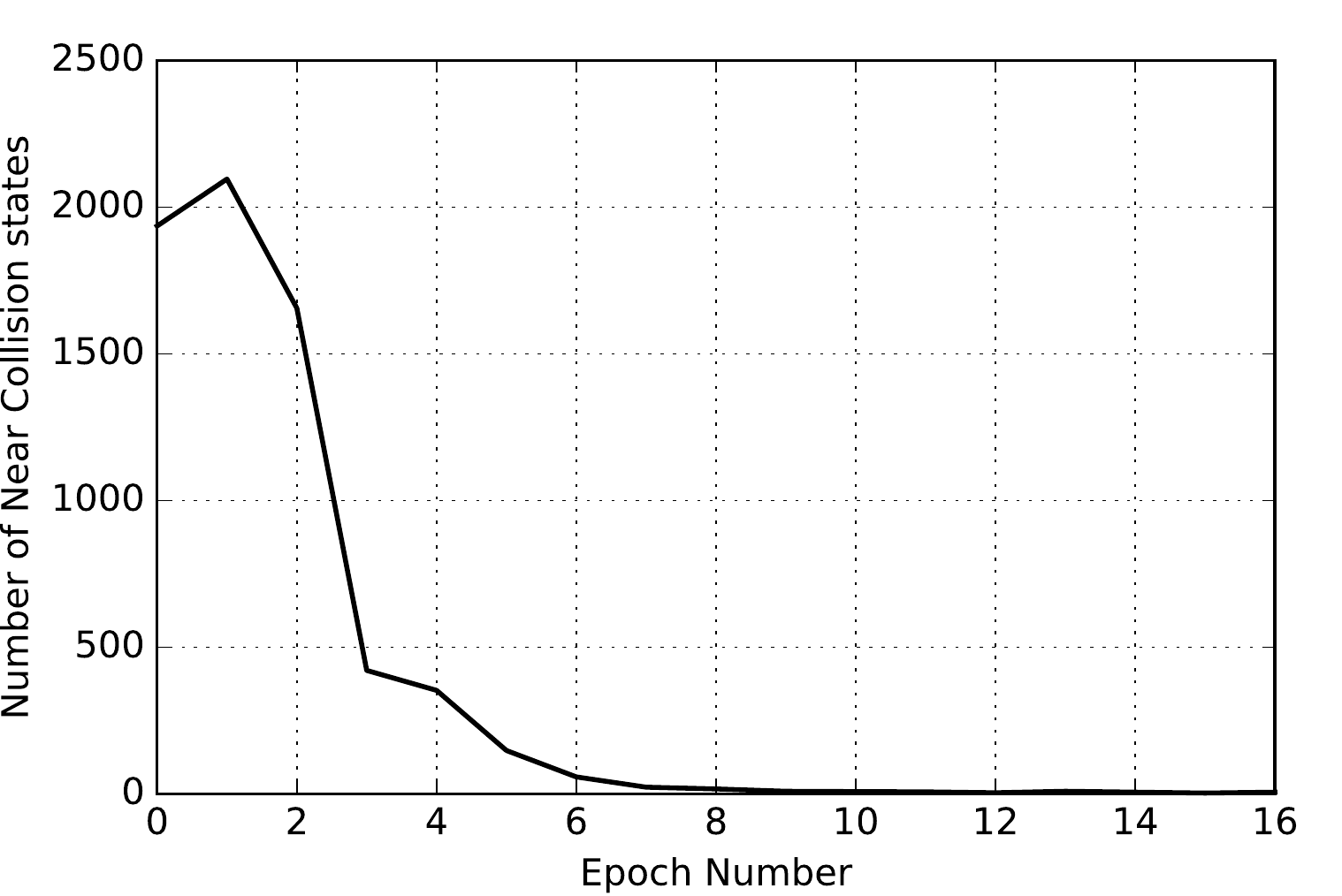}
  \caption{Number of near collision incidents vs training iteration number.}
  \label{fig:collis}
\end{figure}

\begin{figure}[!t]
  \centering
  \includegraphics[width= 0.75 \mycolumnwidth]{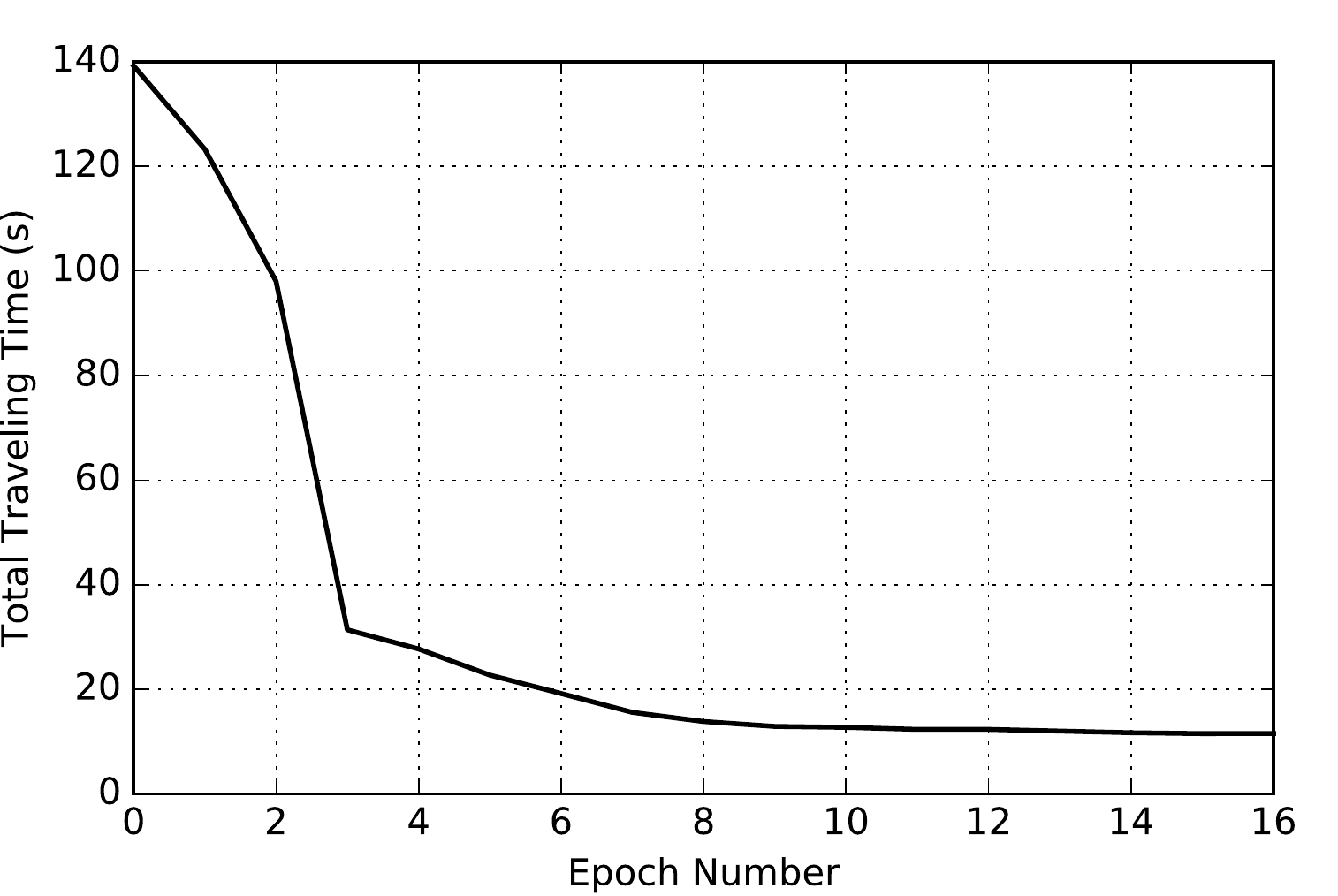}
  \caption{Total travel time of all the vehicles vs training iteration number.}
  \label{fig:time}
\end{figure}

The total number of  ``near collision'' incidents discussed in Section \ref{sec:problem-statement} is shown in Fig. \ref{fig:collis}. Fig. \ref{fig:time} shows the total travel time as a function of training iteration.

\section{Conclusion}
\vspace{-5pt}
In this paper, we have shown that Deep RL can be a promising solution for the problem of intelligent intersection management in local road settings where the number of vehicles is limited and fine-grained acceleration control and motion planning can lead to a more efficient navigation of the autonomous vehicles. We proposed an RL environment definition in which collisions are avoided using a safety mechanism. Using this method instead of large penalties for collision in the reward function, the agent can learn the optimal policy faster and the learned policy can be used in practice where the safety mechanism is actually implemented. The experiments show that the conventional optimization methods are not able to solve the problem with the sizes that are solvable by the proposed method.

Similar to the learning process of human beings, the main benefit of the RL approach is that an explicit mathematical modeling of the system is not required and, more importantly, the challenging task of control design for a complex system is eliminated. However, since the automotive systems demand a high safety requirement, training of the RL agent using a simulation model is inevitable in most cases. However, developing a simulation model for a system is considerably simpler task compared to explicit modeling especially for systems with uncertainty.

While the work at hand is a promising first step towards using RL in autonomous intersection management, a number of potential improvements can be mentioned that might be interesting to address in future work. First, the possibility of developing pre-trained DNNs similar to the works in other mainstream deep learning domains that reduce learning time can be a studied in future. Furthermore, a more advanced rewards system that includes gas usage penalties is another track to developing a practical intelligent intersection management algorithm.

\section{Acknowledgement}
We would like to thank Nvidia for their generous hardware donation. This work was supported in part by the National Science Foundation under NSF grant number 1563652. 

\bibliographystyle{IEEEtran}
\bibliography{IEEEabrv,swc}

\begin{thebibliography}{10}
\providecommand{\url}[1]{#1}
\csname url@samestyle\endcsname
\providecommand{\newblock}{\relax}
\providecommand{\bibinfo}[2]{#2}
\providecommand{\BIBentrySTDinterwordspacing}{\spaceskip=0pt\relax}
\providecommand{\BIBentryALTinterwordstretchfactor}{4}
\providecommand{\BIBentryALTinterwordspacing}{\spaceskip=\fontdimen2\font plus
\BIBentryALTinterwordstretchfactor\fontdimen3\font minus
  \fontdimen4\font\relax}
\providecommand{\BIBforeignlanguage}[2]{{%
\expandafter\ifx\csname l@#1\endcsname\relax
\typeout{** WARNING: IEEEtran.bst: No hyphenation pattern has been}%
\typeout{** loaded for the language `#1'. Using the pattern for}%
\typeout{** the default language instead.}%
\else
\language=\csname l@#1\endcsname
\fi
#2}}
\providecommand{\BIBdecl}{\relax}
\BIBdecl

\bibitem{hausknecht2011autonomous}
M.~Hausknecht, T.-C. Au, and P.~Stone, ``Autonomous intersection management:
  Multi-intersection optimization,'' in \emph{2011 IEEE/RSJ International
  Conference on Intelligent Robots and Systems}.\hskip 1em plus 0.5em minus
  0.4em\relax IEEE, 2011, pp. 4581--4586.

\bibitem{dresner2008multiagent}
K.~Dresner and P.~Stone, ``A multiagent approach to autonomous intersection
  management,'' \emph{Journal of artificial intelligence research}, vol.~31,
  pp. 591--656, 2008.

\bibitem{frese2011comparison}
C.~Frese and J.~Beyerer, ``A comparison of motion planning algorithms for
  cooperative collision avoidance of multiple cognitive automobiles,'' in
  \emph{Intelligent Vehicles Symposium (IV), 2011 IEEE}.\hskip 1em plus 0.5em
  minus 0.4em\relax IEEE, 2011, pp. 1156--1162.

\bibitem{duan2016benchmarking}
Y.~Duan, X.~Chen, R.~Houthooft, J.~Schulman, and P.~Abbeel, ``Benchmarking deep
  reinforcement learning for continuous control,'' \emph{arXiv preprint
  arXiv:1604.06778}, 2016.

\bibitem{schulman2015trust}
J.~Schulman, S.~Levine, P.~Moritz, M.~I. Jordan, and P.~Abbeel, ``Trust region
  policy optimization,'' \emph{CoRR, abs/1502.05477}, 2015.

\bibitem{mladenovic2013self}
M.~N. Mladenovi{\'c} and M.~M. Abbas, ``Self-organizing control framework for
  driverless vehicles,'' in \emph{16th International IEEE Conference on
  Intelligent Transportation Systems (ITSC 2013)}.\hskip 1em plus 0.5em minus
  0.4em\relax IEEE, 2013, pp. 2076--2081.

\bibitem{zohdy2012intersection}
I.~H. Zohdy, R.~K. Kamalanathsharma, and H.~Rakha, ``Intersection management
  for autonomous vehicles using icacc,'' in \emph{2012 15th International IEEE
  Conference on Intelligent Transportation Systems}.\hskip 1em plus 0.5em minus
  0.4em\relax IEEE, 2012, pp. 1109--1114.

\bibitem{malikopoulos2016decentralized}
A.~A. Malikopoulos, C.~G. Cassandras, and Y.~J. Zhang, ``A decentralized
  optimal control framework for connected and automated vehicles at urban
  intersections,'' \emph{arXiv preprint arXiv:1602.03786}, 2016.

\bibitem{sutton1998reinforcement}
R.~S. Sutton and A.~G. Barto, \emph{Reinforcement learning: An
  introduction}.\hskip 1em plus 0.5em minus 0.4em\relax MIT press Cambridge,
  1998, vol.~1, no.~1.

\bibitem{karlik2011performance}
B.~Karlik and A.~V. Olgac, ``Performance analysis of various activation
  functions in generalized mlp architectures of neural networks,''
  \emph{International Journal of Artificial Intelligence and Expert Systems},
  vol.~1, no.~4, pp. 111--122, 2011.

\bibitem{hunting2011aimms}
M.~Hunting, ``The aimms outer approximation algorithm for minlp,''
  \emph{Paragon Decision Technology, Haarlem}, 2011.

\bibitem{schouwenaars2001mixed}
T.~Schouwenaars, B.~De~Moor, E.~Feron, and J.~How, ``Mixed integer programming
  for multi-vehicle path planning,'' in \emph{Control Conference (ECC), 2001
  European}.\hskip 1em plus 0.5em minus 0.4em\relax IEEE, 2001, pp. 2603--2608.

\bibitem{r_openai}
G.~Brockman, V.~Cheung, L.~Pettersson, J.~Schneider, J.~Schulman, J.~Tang, and
  W.~Zaremba, ``Openai gym,'' 2016.

\end{thebibliography}

\end{document}